\pdfoutput=1

\documentclass[11pt]{article}
\usepackage{authblk}
\usepackage{acl}
\usepackage{times}
\usepackage{url}
\usepackage{latexsym}
\usepackage{amssymb}

\usepackage[T1]{fontenc}

\usepackage[utf8]{inputenc}

\usepackage{microtype}
\usepackage{amsmath}
\usepackage{multirow}
\usepackage{booktabs}
\usepackage{graphicx}
\usepackage{float}
\usepackage{hyperref}
\title{Answering Numerical Reasoning Questions in Table-Text Hybrid Contents with Graph-based Encoder and Tree-based Decoder}

\author{\textbf{Fangyu Lei$^{1,2}$,}  \textbf{Shizhu He$^{1,2}$,} \textbf{Xiang Li$^{1,2}$,}  \textbf{Jun Zhao$^{1,2}$,} \textbf{Kang Liu$^{1,2,3}$} \\
$^{1}$National Laboratory of Pattern Recognition, Institute of Automation, CAS, Beijing, China\\
$^{2}$ School of Artificial Intelligence, University of Chinese Academy of Sciences, Beijing, China\\
$^{3}$ Beijing Academy of Artificial Intelligence, Beijing, 100084, China\\

\texttt{\{leifangyu2022, lixiang2022\}@ia.ac.cn}\\
\texttt{\{shizhu.he, jzhao, kliu\}@nlpr.ia.ac.cn}}

\begin{document}
\maketitle
\begin{abstract}
In the real-world question answering scenarios, hybrid form combining both tabular and textual contents has attracted more and more attention, among which numerical reasoning problem is one of the most typical and challenging problems. Existing methods usually adopt encoder-decoder framework to represent hybrid contents and generate answers. However, it can not capture the rich relationship among numerical value, table schema, and text information on the encoder side. The decoder uses a simple predefined operator classifier which is not flexible enough to handle numerical reasoning processes with diverse expressions. To address these problems, this paper proposes a \textbf{Re}lational \textbf{G}raph enhanced \textbf{H}ybrid table-text \textbf{N}umerical reasoning model with \textbf{T}ree decoder (\textbf{RegHNT}). It models the numerical question answering over table-text hybrid contents as an expression tree generation task. Moreover, we propose a novel relational graph modeling method, which models alignment between questions, tables, and paragraphs. We validated our model on the publicly available table-text hybrid QA benchmark (TAT-QA). The proposed RegHNT significantly outperform the baseline model and achieve state-of-the-art results\footnote{We openly released the source code and data at~\url{https://github.com/lfy79001/RegHNT}}~(2022-05-05).

\end{abstract}

\section{Introduction}
Question Answering (QA) is an important task of natural language processing (NLP), which is often used to assess the intelligence of an agent. QA systems use various types of knowledge to answer natural language questions. Earlier approaches independently utilized structured data such as tables~\citep{pasupat2015compositional,yu2018spider}, knowledge bases~\citep{yih2016value, talmor2018web} or unstructured data such as plain texts~\cite{rajpurkar2016squad}. In fact, real-world QA systems often need to fuse different data resources with diverse types in answering complex questions. Therefore, in recent years, the hybrid form of question answering over tables and texts~(TextTableQA) has attracted more and more attention~\citep{chen2020open,chen2020hybridqa, chen2021finqa}.

There are two major question types for TextTableQA. The first is the fact reasoning question, whose answer is usually a span from the table or linked paragraphs, such as the contents in Wikipedia~\citep{chen2020open,chen2020hybridqa}. The second is the numerical reasoning question, which usually aims to use the contents of tables and texts for numerical calculation~\citep{zhu2021tat, chen2021finqa}. Most previous work focuses on the first type, while the numerical reasoning questions have been seldom addressed. The existing datasets such as WikiTableQuestions~\citep{pasupat2015compositional} and DROP~\citep{dua2019drop} also contain numerical reasoning questions, but solving them requires only one type of data source. Therefore, this paper mainly focuses on answering numerical reasoning questions, especially for those complex questions across texts and tables.

\begin{figure*}[h]
    \centering
	\includegraphics[width=\textwidth]{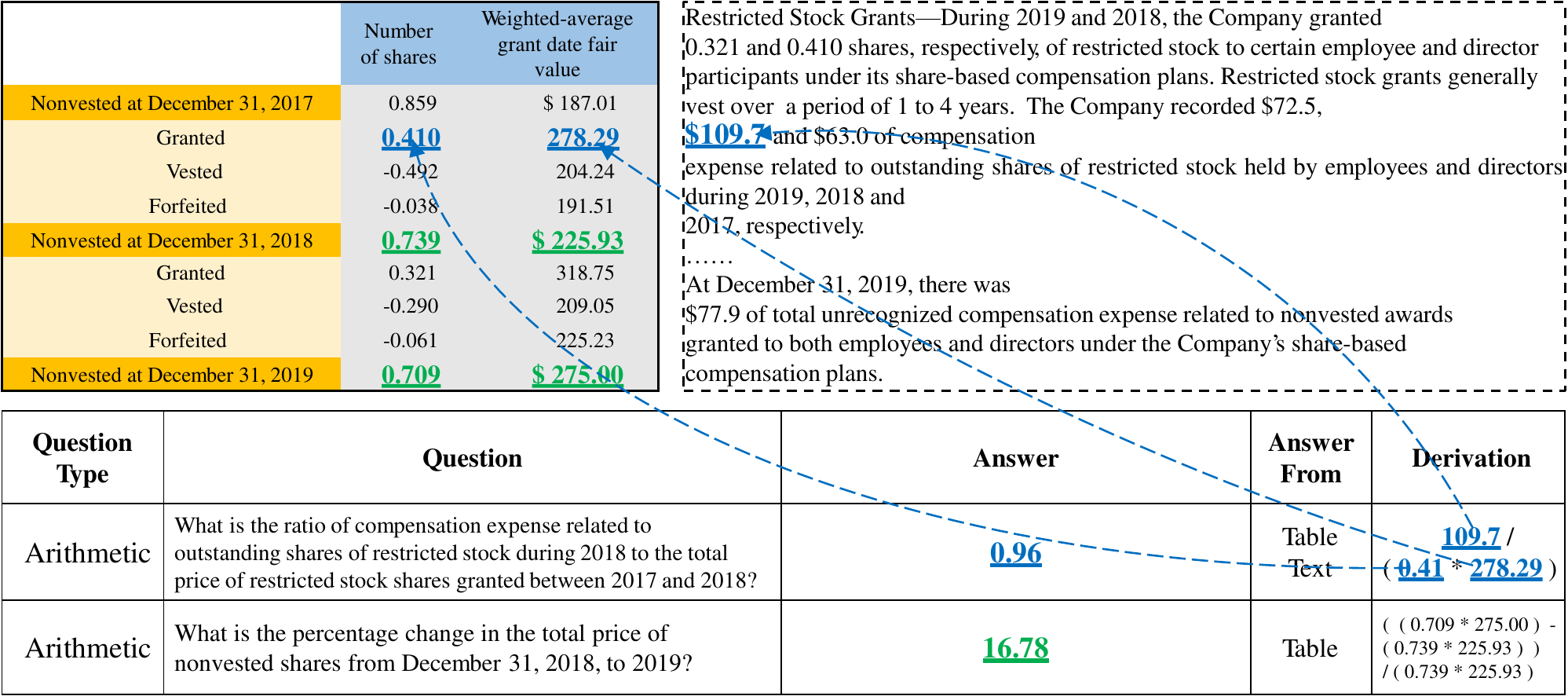}
	\caption{An example of TAT-QA. The solid boxes are tables, and the dotted boxes are the corresponding paragraphs. The bottom table shows two complex questions that cannot be solved by the previous method. The same color marks the source of the answer, while the blue dashed arrow points to the source of the answer. For the table, blue cells are $\mathrm{T_{column}}$, yellow cells are $\mathrm{T_{row}}$ and gold cells are $\mathrm{T_{time}}$.}
	\label{fig1}
\end{figure*}

To explore the application of numerical reasoning questions in hybrid contents. \citet{zhu2021tat} proposed a hybrid text-table dataset TAT-QA, dedicated to fusing the tabular and textual contents to answer numerical reasoning questions. As shown in Figure~\ref{fig1}, for the question ``\emph{What is the ratio of compensation expense related...?}'', one needs to get the numerical value, i.e. ``0.41'', ``278.29'' from the table, and ``109.7'' from the text. Then we need to generate the corresponding numerical expression ``109.7 / ( 0.41 × 278.29 )''. To solve such a numerical question, we need to identify the describing texts near the table and understand the contents of the table and the paragraphs.

Previous method~\citep{zhu2021tat} regarded this problem as a sequence tagging task. They predefine aggregation operators and use a slot filling method to predict simple derivation. An auxiliary number order classifier is used for operators sensitive to the operation orders. Moreover, for the complex numerical computation problems in Figure~\ref{fig1}, the model cannot predict the answer because of the absence of predefined operators. Thus, previous models based on predefined operators have a deficiency of low generalizability and flexibility. In addition, another problem of the method is to model tables and the texts solely and could not aggregate information from different data types. As a result, incorrect answers are usually generated because of the incomprehensive information from a single type of data.

To solve these problems, we propose a novel method to model such table-text hybrid data for the TextTableQA task, especially for the numerical reasoning question type. Specifically, we build a heterogeneous graph to capture the relationship between different data types. And different node types and relation types (intra-relation and inter-relation) are defined, as detailed in Appendix~\ref{appendix:relation}. We expect the QA model to capture the correlation between the tables and the texts and aggregate them effectively. The model could focus on the whole contents rather than a single data type. Then a tree-based decoder is built. And we expect it could make good use of the different data structures from different data types and select appropriate nodes in the heterogeneous graph. After that, an expression tree and a prefix expression are generated. So the model can generate arbitrary forms of derivation without the need for predefined slots. It eliminates error propagation in the operator prediction module and improves the flexibility and generalizability of numerical reasoning.

Experimental results on the public benchmark TAT-QA demonstrate that our proposed model RegHNT improves EM values by 20.2$\%$ and $\mathrm{F_{1}}$ values by 20.0$\%$ over the baseline. Our main contributions are summarized as follows:
\begin{itemize}
\item We design a novel graph construction method to model the information from table-text hybrid data, which effectively captures the correlation between tables and texts.
\item We propose a tree structure decoder to solve the numerical reasoning problems. Based on our method, an expression tree and a prefix expression are generated. Our approach can cover arbitrary numerical derivation forms and improve the model's flexibility and generalizability. To our knowledge, this is the first tree-based model for TextTableQA.
\item We think our graph-tree framework can be used as a strong baseline for the TextTable numerical reasoning task. Empirical results on the TAT-QA dataset demonstrate that the proposed model is effective, which achieves the state-of-the-art performances\footnote{Leaderboard of TAT-QA: \url{https://nextplusplus.github.io/TAT-QA}}.
\end{itemize}

\section{Problem Definition}
\label{section:problem_definition}
We represent a natural language question as $ Q=(q_{1},q_{2},...,q_{|Q|})$ with length $|Q|$. Each question is associated with a table $T$ and a paragraph $P=(s_{1},s_{2},...,s_{|P|})$ with the number of sentences $|P|$. The table $T$ consists of several cells $T=\left \{c_{1},c_{2},... \right \}$ and each table cell $c_{i}$ can be further divided into $K$ words $(c_{i1},c_{i2},...,c_{iK})$. Similarly, each paragraph sentence $s_{i}$ contains several words $(s_{i1},s_{i2},...,s_{iL})$ with the sentence length $L$.
Our goal is to generate the ground truth answer through numerical calculation~(see Figure~\ref{fig1}).

\begin{figure}[h]
    \centering
	\includegraphics[width=0.45\textwidth]{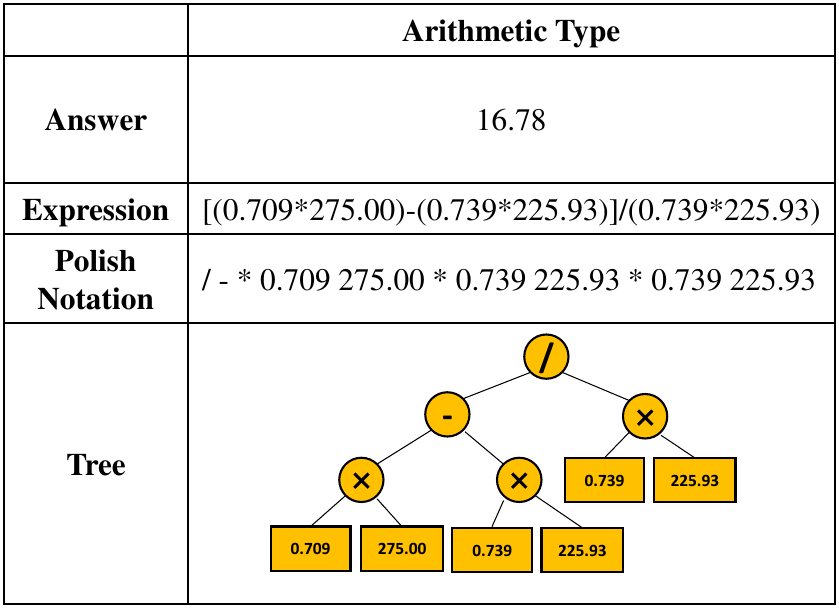}
	\caption{The expressions of arithmetic questions. We predict an mathematical expression consisting of numeric nodes and operators to get the answer.}
	\label{fig3}
\end{figure}

The examples of expressions can be seen in Figure~\ref{fig3}.
For arithmetic questions which need numerical calculation, inspired by math word problem solving~\citep{liu2019tree, wang2019template,zhang2020graph}, we generate the final mathematical expression $\mathrm{E_{p}}$, which is the polish notation transformed from the original math expression. $\mathrm{E_{p}}$ can always be represented as a solution expression tree $\mathrm{T_{e}}$ because the preorder traversal result of the tree is the polish notation.

\begin{figure*}[h]
    \centering
	\includegraphics[width=\textwidth]{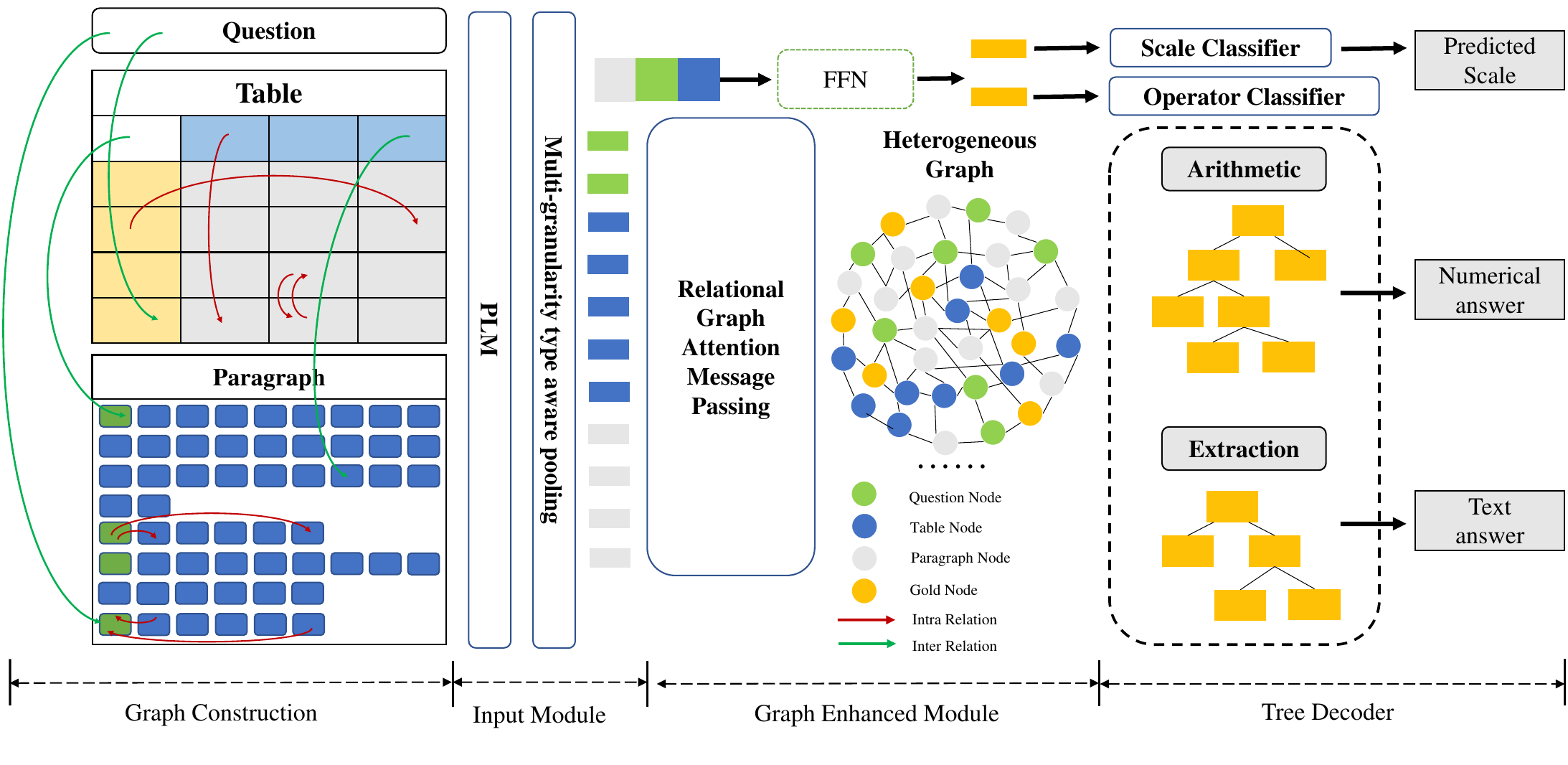}
	\caption{The overall model architecture. The dashed box is the tree-based decoder. Depending on the type of question, two separate trees are constructed to generate the answer.} 
	\label{fig2}
\end{figure*}

\section{Method}
In this section, we present in detail the modules of \textbf{Re}lational \textbf{G}raph enhanced \textbf{H}ybird table-text \textbf{N}umerical reasoning model with \textbf{T}ree decoder~(\textbf{RegHNT})~(see Figure~\ref{fig2}). First, a text-table hybrid data modeling approach is proposed. After constructing the graph, we utilize a classic encoder-decoder architecture for predicting answers. It consists of a graph input module, a graph enhanced hidden module, and a tree-based decoder module. 
Both the graph input module and the graph enhanced hidden module are parts of the encoder, aiming to map the input heterogeneous graph $G$ into node embeddings $\mathbf{Z}  \in  \mathbb{R} ^{|V|\times d}$, where $d$ is the graph hidden size. The tree-based decoder module is responsible for transforming $\mathbf{Z}$ into the target $\mathrm{T_{e}}$.


\subsection{Graph Construction}
The entire input heterogeneous graph $G=(V,R)$ consists of all types of nodes, that is $V=Q\cup T \cup P$ with the number of nodes $|V|=|Q|+ |T| + \sum_{s_{i}\in P}^{}|s_{i}|$, where $|T|$ and $|s_{i}|$ are the number of table cells and paragraph sentences respectively. For relations, $R=R_{I}\cup  R_{C} $. $R_{I}$ denotes intra-relation and $R_{C}$ denotes inter-relation. Details of the node types and relation types are described in the Appendix~\ref{appendix:relation}. Finally, we model the question, table and text as a graph.
\subsection{Graph Input Module}
\label{section:graph_input}
The graph input module aims to initialize embeddings for both nodes and edges. For edges, the edge features are directly initialized from a parameter matrix. For nodes, we can obtain their representations from a pre-trained language model (PLM) such as RoBERTa~\citep{liu2019roberta}.
We flatten all question words, table cells and paragraph words into a sequence $\mathrm{[CLS]}q_{1}q_{2}...q_{|Q|}\mathrm{[SEP]}t_{1}t_{2}...t_{|T|}\mathrm{[SEP]}s_{10}s_{11}...\\
s_{|P|0}s_{|P|1}...\mathrm{[SEP]}$. $s_{i0}$ is the special sentence token of the sentence $s_{i}$.
Since each word $e$ of the sequence is tokenized into sub-words, we use \textbf{Multi-granularity type aware pooling} to get the node representation $\mathrm{x}$. Details in the Appendix~\ref{appendix:subword}.
\subsection{Relational Graph Enhanced Module}
\label{section:graph_hidden}
This module aggregates information about the nodes and edges of the heterogeneous graph. 
It is a stack of $L$ relational graph attention network~(RGAT) layers. 
In each layer $l$, the RGAT (relational graph attention transformers)~\citep{wang2020rat} models the graph $G$ and computes the output representation $\mathbf{Z}$ by:
\begin{equation}
\begin{split}
e_{ij}^{(h)} &=\frac{\mathrm{x}_{i}\mathrm{W}_{q}^{(h)} (\mathrm{x}_{j}\mathrm{W}_{k}^{(h)} +\mathrm{r}_{ij}^{K}  )^{(\mathrm{T})} }{\sqrt{d_{z}/H } } \\
\alpha _{ij}^{(h)} &= \mathrm{softmax}\left \{ e_{ij}^{(h)} \right \} \\
\mathrm{z}_{i}^{(h)}&=\sum_{v_{j}\in \mathcal{N}_{i} }^{}\alpha _{ij}^{(h)}(\mathrm{x}_{j}\mathrm{W}_{v}^{(h)}+\mathrm{r}_{ij}^{V} )
\end{split}
\end{equation}
where matrices $\mathrm{W}_{q}$,$\mathrm{W}_{k}$,$\mathrm{W}_{v}$ are trainable parameters in self-attention, and $\mathcal{N}_{i}$ is the receptive field of node $v_{i}$. The output matrices of the final layer $L$ are the desired outputs of the encoder: $\mathrm{Z} = \mathrm{Z}^{L}  $.

\subsection{Tree-based Decoder Module}
Inspired by the goal-driven tree structure~(GTS)~\citep{xie2019goal} for solving math word problem, we propose a novel tree-based decoder to construct the calculation expressions for solving text-table numerical reasoning problem. As such, the specialized tree decoder generates an equation following the pre-order traversal ordering. The model takes in question $Q$, table $T$, paragraph $P$ and generates a expression tree $\mathrm{T_{e}}$. Let $V_{num}$ denote numeric values in $T$ and $P$. Generally, $V_{con}$ denotes constant values $V_{con}=\left \{1,\mathrm{AVG}\right \}$, $\mathrm{AVG}$ means to average the sum of the previous numbers. $V_{op}$ denotes mathematical operators $V^{op}=\left \{ +,-,\times,{\div} \right \} $.

The tree generation process is designed as a preorder tree traversal~(root-left-right). For node $y$ in target $\mathrm{T_{e}}$, $y\in V^{num}\cup V^{con}\cup V^{op}$. We set $V_{num}$ and $V_{con}$ to be the leaf nodes and $V_{op}$ serve as the internal nodes and must have two child nodes.

The tree structured decoder uses the final graph layer representations $\mathrm{z}_{i}$ as input and generates the target expression in $\mathrm{t}$ time steps. At each time step $\mathrm{t}$, let $\mathrm{s_{t}}$ denote the decoding hidden state, $\mathrm{c_{t}}$ denotes the hybrid \textbf{c}ontext state, $\mathrm{g_{t}}$ denotes the \textbf{g}enerated expressions tree state.   

The decoder is a  bi-directional GRU~\citep{cho2014learning}, which updates its states at time step $\mathrm{t}+1$ as follows:
$$\mathrm{s}_{\mathrm{t}+1}=\mathrm{BiGRU}([\mathrm{c}_{\mathrm{t}}:g_{\mathrm {t}}:\mathrm{E}(y_{t}) ],\mathrm{s}_{\mathrm{t}})$$
where $\mathrm{E}(y_{t})$ is the embedding of token $\mathrm{y_{t}}$:
\begin{equation}
\begin{split}
\mathrm{E}(y_{t})=\left\{
\begin{aligned}
&\mathbf{M}_{op}(y_{t})\quad \mathbf{if\enspace y_{t}}\in V^{op}  \\
&\mathbf{M}_{con}(y_{t})\quad \mathbf{if\enspace y_{t}}\in V^{con}  \\
&h_{loc(y_{t},T,P)}^{i}\quad \mathbf{if\enspace y_{t}}\in V^{num} 
\end{aligned}
\right.
\end{split}
\end{equation}
$\mathbf{M}_{op}$ and $\mathbf{M}_{con}$ are two trainable embeddings for operators and constants, respectively. For a numeric value in $V^{num}$, its token embedding takes the corresponding hidden state $h_{loc(y_{t},T,P)}^{i}$, where $loc(y_{t},T,P)$ is the index position of $y$ in table $T$ or paragraph $P$~\cite{hong2021learning}.

Inspired by math word problem solving~\citep{wu2021edge}, the generated expression tree state $\mathrm{g_{t}}$ is calculated as follows:
\begin{equation}
\mathbf{\mathrm{g_{t+1}}} = \sigma(\mathrm{W_{g}}[\mathrm{g_{t}:g_{g,p}:g_{t,l}:g_{t,r}}])
\end{equation}
where $\sigma$ is a sigmoid function and $\mathrm{W_{g}}$ is a weight matrix. For each generated node, $\mathrm{g_{g,p}}$, $\mathrm{g_{g,l}}$, $\mathrm{g_{g,r}}$ represent the expression tree state of the parent node, left child node, and right child node of the current node, respectively.

The hybrid context state $\mathrm{c}_{\mathrm{t}}$ is computed via attention mechanism as follows:
\begin{equation}
\begin{split}
\alpha _{ti}=\mathrm{softmax}(&\mathrm{tanh}\mathrm{W_{h}}\mathbf{z_{i}}+\mathrm{W_{s}}[\mathbf{s_{t}} :\mathbf{r_{t}} ]  ))\\
\mathbf{c_{t}}&=\sum_{i=1}^{m}\alpha _{ti}\mathbf{z_{i}}
\end{split}
\end{equation}
where $\mathrm{W_{h}}$, $\mathrm{W_{s}}$ are weight matrices. $\alpha _{ti}$ is the attention distribution on the node representations $\mathbf{z_{i}}$.

Lastly, the decoder can generate a word from a given vocabulary $V_{op}\cup V_{con}$. It can also generate a number symbol from $V_{num}$, which is copied a number from the table $T$ or paragraph $P$. The final distribution is the combination of the generated probability and copy probability:
\begin{equation}
\begin{split}
p_{c}&=\sigma(\mathrm{W_{z}} [\mathbf{s_{t}:c_{t}:r_{t}} ] )\\
\mathrm{P_{c}}(y_{t} ) &=\sum_{y_{t}=\mathbf{x_{i}}}^{} \alpha _{ti}\\
\mathrm{P_{g}}(y_{t} ) &= \mathrm{softmax}(f([\mathbf{s_{t}:c_{t}:r_{t}}  ]))\\
\mathrm{P} (y_{t}|y_{<t},\mathrm{X} )&=p_{c} \mathrm{P_{c}} (y_{t}) + (1-p_{c})\mathrm{P_{g}} (y_{t})
\end{split}
\end{equation}
Here, $f(\cdot )$ is a perception layer. $p_{c}$ is the probability that the current word is a number copied from the table or paragraphs.

\subsection{Operator and Scale Prediction}
\label{sec:opertor_and_scale}
In addition, there are two separate tasks in the decoding section: \textbf{operator prediction} and \textbf{scale prediction}. For arithmetic questions, a right prediction of a numerical answer should include the right number and the correct scale. The scale in the dataset may be \emph{None}, \emph{Thousand}, \emph{Million}, \emph{Billion}, and \emph{Percent} generally. We focus on arithmetic questions for operator prediction, but there are still non-arithmetic questions~(span extraction question) in the dataset. So we classify whether the question is arithmetic or not before decoding. For the extraction questions, as shown in Figure~\ref{fig2}, we also model them as trees, as described in Appendix~\ref{appendix:extraction}.

To predict the right aggregation operator and scale, two multi-class classifiers are developed. In particular, we take the vector $\mathrm{\left \langle CLS \right \rangle}$ to compute the probability:
\begin{equation}
\begin{split}
\mathrm{p^{op}} &=\mathrm{softmax(FFN}   (    \left [  \mathrm{\left \langle CLS \right \rangle};h_{Q};h_{T};h_{P} \right ] ))\\
\mathrm{p^{scale}} &=\mathrm{softmax(FFN}   (    \left [  \mathrm{\left \langle CLS \right \rangle} \right ] ))
\end{split}
\end{equation}
where $h_{Q}$,$h_{T}$ and $h_{P}$ are the representations of the question, the table and the paragraphs , respectively, which are obtained by applying an average pooling over the representations of their corresponding tokens. “;” denotes concatenation, and FFN denotes a two-layer feed-forward network with the GELU activation.
\subsection{Training}
To optimize \textbf{RegHNT}, the overall loss is the sum of the loss of the above tasks:
\begin{equation}
\begin{split}
\mathcal{L} &= \mathcal{L}_{tree}+ \mathcal{L}_{op} + \mathcal{L}_{scale}\\
\mathcal{L}_{tree}&=-\sum_{t=1}^{T}\mathrm{log} \mathbf{P} (y_{t}|y_{<t},\mathrm{Q,T,P})\\
\mathcal{L}_{op}&=\mathrm{NLL(log(P^{op}), G^{op})}\\
\mathcal{L}_{scale}&=\mathrm{NLL(log(P^{scale}), G^{scale})}
\end{split}
\end{equation}
$\mathcal{L}_{tree}$ is the loss function of training the tree-decoder, and we use the cross-entropy loss. $\mathcal{L}_{op}$ and $\mathcal{L}_{scale}$ are the loss functions for operator prediction and scale prediction, respectively, where NLL(·) is the negative log-likelihood loss. $\mathrm{G^{op}}$ comes from the supporting evidence, which is extracted from the annotated answer and derivation. $\mathrm{G^{scale}}$ uses the annotated scale of the answer. We add up the three loss functions as the total loss function.

\section{Experiments}

\subsection{Dataset and Evaluation Metrics}
TAT-QA~\citep{zhu2021tat} is a large-scale, hybrid QA dataset which contains numerical reasoning and span extraction questions. And the contents of TAT-QA include both tabular and textual data from real financial reports. It contains a total of 2,757 hybrid contexts and 16,552 corresponding question-answer pairs. The detailed statistics are shown in Appendix~\ref{appendix:tatqa}. The original dataset contains four types of questions: \emph{Span}, \emph{Multi-Span}, \emph{Count}, \emph{Arithmetic}. But in our setup, there are two types of questions: \emph{Span Extraction} and \emph{Arithmetic}. And our work mainly focuses on answering the \emph{Arithmetic} questions.

For evaluation, we adopt the Exact Match (EM) and numeracy-focused F1 score~\citep{dua2019drop} to measure the performance of different QA models. All of which are computed using the official evaluation script\footnote{\url{https://github.com/NExTplusplus/tat-qa}}. We submit our model to the organizer of the challenge for evaluation. The evaluation detail can be found on the original paper~\citep{zhu2021tat}.

\subsection{Implementation Details}
\textbf{Implementations.} Our model is implemented with PyTorch~\citep{paszke2019pytorch}, and the graphs are constructed with the library DGL~\citep{wang2019deep}. In the graph input module, we use pre-trained language models (PLMs) RoBERTa~\citep{liu2019roberta} to obtain the initial representations. During evaluation, we adopt beam search decoding with beam size 3. \\
\textbf{Hyper-parameters.}
In the encoder, the number of GNN layers $L$ is 8, and the number of heads in multi-head attention is 8. For PLMs, we use learning rate 1e-5 and weight decay rate 0.01. For other model modules, we use a larger learning rate 1e-4, and a weight decay rate 5e-5. In the decoder, The recurrent dropout rate~\citep{gal2016theoretically} is 0.2 for GRU. The number of heads in multi-head attention is 8 and the dropout rate of features is set to 0.1 in both the encoder and decoder. Throughout the experiments, we use AdamW~\citep{loshchilov2018decoupled} optimizer with a linear warmup scheduler. The warmup ratio of the total training steps is 0.06. The batch size is 48, and the training epoch is 100. The training process may take around 2 days using a single NVIDIA GeForce RTX 3090.\\
\textbf{Baselines.} We compare with the standard T{\small AG}O{\small P}~\citep{zhu2021tat}, which first applies sequence tagging to extract relevant cells from the table and text spans from the
paragraphs. We also compare with other advanced models, which can be found on the TAT-QA challenge leaderboard\footnote{\url{https://nextplusplus.github.io/TAT-QA}}. There are no linked papers to the submissions as yet. We compare our model’s performance on the test split
with all of them.
\subsection{Main Results}

\begin{table}[h]\small
\centering
\begin{tabular}{@{}ccccc@{}}
\toprule
\multirow{2}{*}{\textbf{Method}} & \multicolumn{2}{c}{\textbf{Dev}} & \multicolumn{2}{c}{\textbf{Test}} \\ \cmidrule(l){2-5} 
                                 & EM              & $\mathrm{F_{1}}$  & EM           & $\mathrm{F_{1}}$             \\ \midrule
Human                            & -               & -              & 84.1            & 90.8            \\ \midrule
T{\scriptsize  AG}O{\scriptsize  P}                            & 55.2            & 62.7           & 50.1            & 58.0            \\ \midrule
LETTER                           & -               & -              & 56.1            & 64.3            \\ \midrule
KIQA                             & -               & -              & 58.2            & 67.4            \\ \midrule
UniRPG~\cite{zhou2022unirpg}                       & 70.2            & 77.9              & 67.4            & 75.5            \\ \midrule
\textbf{RegHNT}                  & \textbf{73.6}   & \textbf{81.3}  & \textbf{70.3}   & \textbf{78.0}   \\ \bottomrule
\end{tabular}
\caption{The performance of different models on dev and test set of TAT-QA. The best results are marked in bold.}
\label{tab:main_result}
\end{table}

\begin{table}[h]\small
\centering
\begin{tabular}{@{}ccccc@{}}
\toprule
           & Table     & Text      & Table-text      \\ \cmidrule(l){2-4} 
           & EM/$\mathrm{F_{1}}$     & EM/$\mathrm{F_{1}}$     & EM/$\mathrm{F_{1}}$\\ \midrule
\multicolumn{4}{c}{T{\scriptsize  AG}O{\scriptsize  P}}                                   \\ \midrule
Span       & 56.5/57.8 & 56.5/57.8 & 68.2/71.7   \\
Spans      & 66.3/77.0 & 19.0/59.1 & 63.2/76.9   \\
Counting   & 63.6/63.6 & -/-       & 62.1/62.1   \\
Arithmetic & 41.1/41.1 & 27.3/27.3 & 46.5/46.5   \\
 \midrule
\multicolumn{4}{c}{RegHNT}                                  \\ \midrule
Span       & \textbf{68.5}/\textbf{70.0} & \textbf{58.7}/\textbf{83.0} & \textbf{77.0}/\textbf{84.7}   \\
Spans      & \textbf{79.5}/\textbf{86.2} & \textbf{23.8}/\textbf{65.3} & \textbf{81.1}/\textbf{90.1} \\
Counting   & 36.3/36.3 & -/-       & \textbf{82.7}/\textbf{82.7}   \\
Arithmetic & \textbf{72.7}/\textbf{72.7} & 27.3/27.3 & \textbf{77.7}/\textbf{77.7}   \\
 \bottomrule
\end{tabular}
\caption{Detailed experimental results of T{\small AG}O{\small P} and RegHNT w.r.t. answer types and sources on the test set.}
\label{tab:detail_result}
\end{table}

The main results on the test set are provided in Table~\ref{tab:main_result}. Our model achieves the state-of-the-art results in the publicly available TAT-QA benchmark and achieves 20$\%$ higher on both EM and $\mathrm{F_{1}}$ compared with the original baseline (T{\small AG}O{\small P}), which shows that our model can answer more questions with higher accuracy.

The detailed results on the test set are provided in Table~\ref{tab:detail_result}. For almost all types of questions, the accuracy of RegHNT prediction has been improved. Thanks to the tree decoder for arithmetic questions, the model's accuracy on this type of questions has been greatly improved, with an overall improvement of almost 30$\%$. For span extraction questions~(\emph{Span}, \emph{Spans}, \emph{Counting}), we observe a improvement in the performance of the model as well. From the perspective of answer sources, compared with the original baseline (T{\small AG}O{\small P}), we focus on "table-text" type questions, both "arithmetic" and "span extraction" type show a performance improvement of about 20$\%$. It demonstrates that our approach works very well in solving table-text hybrid questions.

This paper focuses on the numerical reasoning questions. To verify our model more precisely, we divided the questions into three categories based on the complexity of the derivation, as shown in Table~\ref{tab:ari_study}. Simple arithmetic has only one operator. Complex arithmetic has multiple operators, usually used to calculate the average and the rate of change. Undefined arithmetic is arithmetic for which no template is defined in T{\small AG}O{\small P}. The results show that our model significantly improves both simple and complex arithmetic. In particular, our model can solve undefined arithmetic of T{\small AG}O{\small P}, which offers flexibility and generalizability.

\begin{table}[h]\small
\centering
\begin{tabular}{@{}lccc@{}}
\toprule
Arithmetic type      & \%   & T{\scriptsize AG}O{\scriptsize P} & RegHNT        \\ \midrule
Simple arithmetic    & 41.8 & 45.0  & \textbf{79.3} \\ \midrule
Complex arithmetic   & 42.8 & 60.5  & \textbf{85.3} \\ \midrule
Undefined arithmetic & 15.4 & 0.0   & \textbf{61.8} \\ \bottomrule
\end{tabular}
\caption{Exact match value for different types of arithmetic questions.}
\label{tab:ari_study}
\end{table}

\begin{table}[h]\small
\centering
\begin{tabular}{lcc}
\hline
Technique           & EM   &$\mathrm{F_{1}}$   \\ \hline
RegHNT              & \textbf{73.6} & \textbf{81.3} \\
w/o Intra-relations & 72.8 & 80.4    \\
w/o Inter-relations & 72.3 & 79.8    \\
w/o All relations   & 71.6 & 78.7    \\ 
w/o Multi-granularity type aware pooling   & 73.0 & 80.7    \\
w/o Tree-decoder~(Span extraction)  & 72.8 & 80.6    \\ 
w/o Tree-decoder~(All questions)            & 60.3 & 69.5    \\ \hline
\end{tabular}
\caption{Ablation study of different modules.}
\label{tab:ablation_study}
\end{table}

\subsection{Ablation Studies}
\label{section:ablation}
\textbf{Effect of Tree Decoder}. Although our work focuses on arithmetic questions, to unify the whole model into a graph-tree framework, we transform the span extraction type question into a tree generation problem as well, as mentioned in Section~\ref{sec:opertor_and_scale}. We conducted two experiments using the sequence tagging method in T{\small AG}O{\small P}~\citep{zhu2021tat} instead of generating expression trees. As shown in the last two rows of Table~\ref{tab:ablation_study}, one is to use sequence tagging for all questions, and the other is to use sequence tagging only for span extraction questions. When we change the decoder only for extraction questions, the $\mathrm{F_{1}}$ drops only 0.7$\%$, but when we change the decoder for all questions, the $\mathrm{F_{1}}$ drops about 11.8$\%$. It shows that the tree decoder is not only tremendously helpful in solving arithmetic questions but also provides a slight improvement in solving span extraction questions.\\
\textbf{Effect of Graph Encoder.} As shown in the first four rows of Table~\ref{tab:ablation_study}, we show the effectiveness of the proposed graph encoder. Removing the intra-relations reduces the $\mathrm{F_{1}}$ value by 0.9$\%$ and reduces the EM value by 0.8$\%$. Removing the inter-relations reduces $\mathrm{F_{1}}$ value by 1.5$\%$ and reduces the EM value by 1.3$\%$. When we remove all relations~(remove graph enhanced module), the $\mathrm{F_{1}}$ decreases by 2.6$\%$, and the EM value decreases by 2.0$\%$. From the results, it can be clearly confirmed that the graph we built plays an essential role in modeling table-text hybrid data, and it captures the semantic association through the message passing between different data types.\\
\textbf{Effect of Subword Pooling Layer.} In the graph input module, we used a multi-granularity type aware pooling method. The type classification criteria for word granularity are text and number, and for node granularity are question, table, and paragraph. As shown in the fifth row ``w/o Multi-granularity type aware pooling'' of Table~\ref{tab:ablation_study}, we eliminate this mechanism and unify the pooling approach for all types and granularities. Experimental results of $\mathrm{F_{1}}$ dropped by 0.6$\%$, which shows the effectiveness of this type-aware module.
\subsection{Scale and Operater Study}
\textbf{Scale Study.} Scale prediction is a unique challenge over TAT-QA and very pervasive in the context of finance. After obtaining the scale, the numerical or string prediction is multiplied or concatenated with the corresponding scale as the final prediction to compare with the ground-truth answer, respectively. We compare RegHNT with the baseline model for scale prediction results. The experimental results are shown in Table~\ref{tab:scale}. Our model has significantly improved performance on both the dev and test datasets. To explore the impact of the scale on results, we use the gold scale to predict the answer. As shown in the third row of Table~\ref{tab:oracle_study}, model accuracy will slightly increase to 84.2$\%$ when we use the gold scale, which shows that it is necessary to improve the prediction of scale.\\
\textbf{Operator Study.} For TAT-QA dataset, there are four original answer types: \emph{Span}, \emph{Multi-Span}, \emph{Count}, \emph{Arithmetic}. As Figure~\ref{fig2} shows, we have adapted it into two categories, where the details of the expression tree for the span extraction questions are in the Appendix~\ref{appendix:extraction}. To investigate whether this category setting causes error propagation, we use the gold operator to predict the answer, and the results are shown in Table~\ref{tab:oracle_study}. When we use the gold operator, the EM and $\mathrm{F_{1}}$ of the model is improved by only 0.1. It suggests, to some extent, that we divide the data into two categories and use tree decoders to generate the answers separately. This approach has no significant impact on performance.
\begin{table}[h]\small
\centering
\begin{tabular}{@{}ccc@{}}
\toprule
Model  & Dev  & Test \\ \midrule
T{\scriptsize  AG}O{\scriptsize  P} & 93.5 & 92.2 \\ \midrule
RegHNT  & \textbf{95.3} & \textbf{93.4}    \\ \bottomrule
\end{tabular}
\caption{Scale prediction results of our model and baseline.}
\label{tab:scale}
\end{table}

\begin{table}[h]\small
\centering
\begin{tabular}{@{}lcc@{}}
\toprule
Model                                   & EM   &$\mathrm{F_{1}}$ \\ \midrule
RegHNT                                  & 73.6   & 81.4\\
RegHNT + Gold operator                & 73.7   & 81.5 \\
RegHNT + Gold scale                   & 76.5   & 84.2   \\
RegHNT + Gold operator + Gold scale & \textbf{76.7} & \textbf{84.3}     \\ \bottomrule
\end{tabular}
\caption{The performance of using gold operators and gold scales.}
\label{tab:oracle_study}
\end{table}

\subsection{Case Studies}

\begin{figure}[h]
    \centering
	\includegraphics[width=0.45\textwidth]{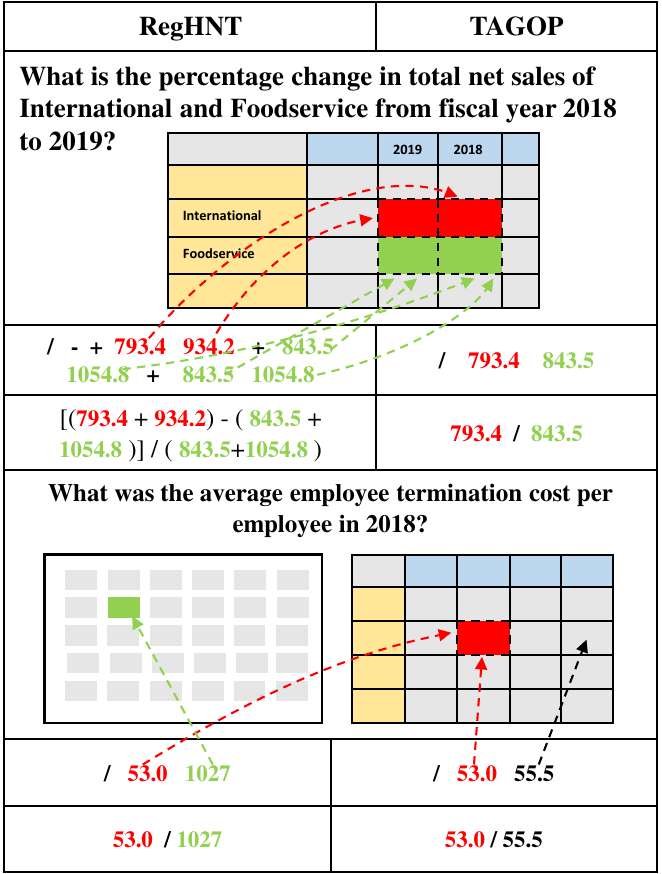}
	\caption{Two examples of generated expressions by RegHNT and T{\small AG}O{\small P}\citep{zhu2021tat}.}
	\label{fig4}
\end{figure}

As Figure~\ref{fig4} shows, there are two questions and the prediction results of the models. For the question ``\emph{What is the percentage change in total net sales...?}'', the previous method could not generate complex arithmetic. In contrast, our model can generate the correct prefix expressions, which demonstrates the characteristics and advantages of our tree decoder. For the question ``\emph{What was the average employee termination...?}", the correct expression is derived from table and text~(``53.0'' from table and ``1027'' from paragraph). The results in Figure~\ref{fig4} show that the model often fails to answer correctly when a question requires the use of both tables and text. It focuses only on tables, choose ``53.0'' and ``55.5''. This error type is very common in the previous methods. It shows that our graph encoder can better model the association between tables and texts.
\section{Related Work}
Table-text hybrid QA is a new task that consists of two main types of work.\\

\textbf{Fact Reasoning TextTableQA.}~\citet{chen2020hybridqa} propose the first table-text hybrid QA dataset. It is the fact reasoning type dataset whose answer is usually a span from the table or linked paragraphs of Wikipedia. The authors supposed HYBRIDER~\citep{chen2020hybridqa}, a pipeline approach that divides the prediction process into two phases called linking and reasoning. MITQA~\citep{kumar2021multi} achieves SOTA EM result on HybridQA, which is a novel training strategy that works with multiple instances and multiple answers based on weak supervision. DEHG~\citep{feng-etal-2022-multi} propose a document-entity heterogeneous graph network and achieve SOTA F1 score on HybridQA. OTT-QA~\citep{chen2020open} is a difficult open-domain setting TextTableQA task, which needs retrieval and reading to get the answers. CARP~\citep{ijcai2022p629} utilizes hybrid chain to model the explicit intermediate reasoning process across table and text for question answering, which achieves SOTA results, but still far from expectations. GeoTSQA~\cite{li2021tsqa} is a multiple choice QA dataset based on geography domain.\\

\textbf{Numerical Reasoning TextTableQA.}
TAT-QA~\citep{zhu2021tat} and FinQA~\citep{chen2021finqa} are the numerical reasoning hybrid dataset which comes from the financial field. Both TAT-HQA~\citep{li2022learning} and TAT-DQA~\citep{zhu2022towards} are enhanced datasets of TAT-QA, which study TextTableQA in counterfactual condition and multimodal condition, respectively. M{\small ULTI}H{\small IERTT}~\citep{zhao-etal-2022-multihiertt} is a challenging dataset, which contains multiple hierarchical tables and longer unstructured text.
Unlike HybridQA, which was fact reasoning type questions, TAT-QA focuses explicitly on finance and needs numerical reasoning for question answering over tabular numbers and associated text. They proposed T{\small AG}O{\small P} model and regarded this problem as a sequence tagging task. It predefined aggregation operators and used a slot filling method to predict simple derivation, lacking generalizability and flexibility. Our model is the first method to generate arithmetic expressions directly for table-text hybrid numerical reasoning QA.

\section{Conclusion}
This paper proposes a novel method to solve table-text hybrid numerical reasoning problems and achieve good performance. We present a unified framework for addressing the table schema and relative paragraphs. By adopting relation-aware self-attention, the proposed method jointly learns question, table and paragraph representations based on their alignment. At the same time, we offer a tree-based numerical reasoning decoding framework for hybrid data, the first to use this type of method for this task. Our model can serve as a strong baseline for this task. However, our model has not yet been experimented on encyclopedic type questions~\citep{chen2020hybridqa}, and we will explore a table-text hybrid QA framework that integrates dealing with factual and numerical reasoning types of questions. For numerical reasoning type questions, we will do further research on TAT-HQA~\citep{li2022learning} and TAT-DQA~\citep{zhu2022towards}.

\section*{Acknowledgements}

This work is supported by the Natural Key R\&D Program of China (No.2022QY0701), the National Natural Science Foundation of China (No.61922085, No.61976211) and the Strategic Priority Research Program of Chinese Academy of Sciences (Grant No. XDA27020200). This research work was supported by the independent research project of National Laboratory of Pattern Recognition (No. Z-2018013), the Youth Innovation Promotion Association CAS, Yunnan Provincial Major Science and Technology Special Plan Projects (No.202202AD080004) and CCF-DiDi GAIA Collaborative Research Funds for Young Scholars.

\bibliography{custom}
\bibliographystyle{acl_natbib}

\appendix
\section{Relation Detail}
\label{appendix:relation}

Before defining relation types, we'd better introduce the node types more finely. There are three types of nodes, but they can be divided in more detail. Table $T$ has complex structural information. We set three types according to the location of the cell, namely $\mathbf{T_{row}}$, $\mathbf{T_{column}}$ and $\mathbf{T_{cell}}$. Some row headers and column headers are time~(e.g. 2019). We set these special cell as $\mathbf{T_{time}}$. Row headers and column headers are composed of text, while numerical cells are numbers. For the question $Q$, there is only one node type $\mathbf{Q_{word}}$. For the paragraph $P$, there are two node types, $\mathbf{P_{word}}$ represents the common sentence words, $\mathbf{P_{sentence}}$ represents the special sentence token.\\
Now there are seven types of nodes, and we establish edges between them, which is mainly divided into intra-relation $\mathrm{R_{I}}$ and inter-relation $\mathrm{R_{C}}$. The relations are based on the table schema and word matching between resources, and details are described in Table~\ref{tab:relation}.

\begin{table*}[h]
\centering
\begin{tabular}{@{}cccl@{}}
\toprule
\multicolumn{4}{c}{Intra-Relation}                                                                                              \\ \midrule
Source $x$                 & Source $y$                & Relation     & Description                                                 \\ \midrule
$\mathrm{T_{row}}$  & $\mathrm{T_{cell}}$  & CONTAIN    &$x$ is row head of $y$.                                          \\ \midrule
$\mathrm{T_{column}}$     & $\mathrm{T_{cell}}$   & CONTAIN   &$x$ is column head of $y$.                                       \\ \midrule
$\mathrm{T_{time}}$       & $\mathrm{T_{cell}}$    & CONTAIN   &$x$ is row/column head of $y$, $x$ is time.                        \\ \midrule
$\mathrm{P_{sentence}}$  &$\mathrm{P_{word}}$     & CONTAIN     &$x$ is the sentence token which contain $y$.                     \\ \midrule
$\mathrm{Q_{word}}$     & $\mathrm{Q_{word}}$   & DISTANCE+1   & $y$ is the next word of $x$.                                    \\ \midrule
$\mathrm{P_{word}}$     & $\mathrm{P_{word}}$        & DISTANCE+1   & $y$ is the next word of $x$.                                    \\ \midrule
$\mathrm{T_{cell}}$         &$\mathrm{T_{cell}}$ &SAME ROW    & $x$ and $y$ are in the same row.                                \\ \midrule
$\mathrm{T_{cell}}$         &$\mathrm{T_{cell}}$ &SAME COLUMN    & $x$ and $y$ are in the same column.                                \\ \midrule
\multicolumn{4}{c}{Inter-Relation}                                                                                              \\ \midrule
\multirow{2}{*}{$\mathrm{Q_{Word}}$} & \multirow{2}{*}{$\mathrm{T_{row}}$ } & PARTIALMATCH &$x$ is part of $y$, but the entire question does not contain $y$. \\
     &    & EXACTMATCH  &$x$ is part of $y$, and $y$ is a span of the entire question.     \\ \midrule
$\mathrm{P_{sentence}}$     & $\mathrm{Q_{word}}$        & CONTAIN      &$x$ is the sentence token which contain $y$.                    \\ \midrule
$\mathrm{P_{sentence}}$      & $\mathrm{T_{row}}$      & CONTAIN   &$x$ is the sentence token which contain $y$.                    \\ \midrule
\multirow{2}{*}{$\mathrm{P_{word}}$ } & \multirow{2}{*}{$\mathrm{T_{row}}$ }  & PARTIALMATCH &$x$ is part of $y$, but the entire sentence does not contain $y$. \\
                         &              & EXACTMATCH  &$x$ is part of $y$, and $y$ is a span of the entire question. \\ \midrule
$\mathrm{P_{word}}$         & $\mathrm{Q_{word}}$         &SAME    &$x$ and $y$ are the same words.                                 \\ \bottomrule
\end{tabular}
\caption{The checklist of all relations in our RegHNT. All the above relations are asymmetric. We show only one direction, and the opposite direction can be easily inferred.}
\label{tab:relation}
\end{table*}

\section{Multi-granularity type aware pooling}
\label{appendix:subword}
Since each word $e$ of the sequence is tokenized into sub-words, we need to aggregate them in order to obtain the node representation. We set two granularities ~(word, node) and three types~(question, table, paragraph) aware pooling method. 
\begin{itemize}
\item Word level: For number word~(e.g. 109.7) and text word~(e.g. compensation), we use two independent subword attentive pooling module depending on word type to get the type-aware word level representation $\mathrm{w}$.
$$\mathrm{w}=\sum_{i}^{} \mathrm{softmax}_{i} \left [ \mathrm{tanh}(\mathrm{e}_{i}\mathrm{W}_{s})\mathrm{v}_{s}^\mathrm{T} \right ]\mathrm{e}_{i}$$
\item Node Level: For a node, especially a table cell node, which usually consists of multiple words. According to the node source types, we aggregate the word level granularity representation with three different pooling layers and three different BiLSTM to get the node representation $\mathrm{x}$.
$$\mathrm{x}=\sum_{i}^{} \mathrm{softmax}_{i} \left [ \mathrm{tanh}(\mathrm{w}_{i} \mathrm{W}_{n})\mathrm{v}_{n}^\mathrm{T} \right ]\mathrm{w}_{i}$$
\end{itemize}
Where $\mathrm{v}_{s}$, $\mathrm{W}_{s}$, $\mathrm{v}_{n}$, $\mathrm{W}_{n}$ are trainable parameters. The attentive pooling layer is inspired by LGESQL~\citep{cao2021lgesql}.

\section{Details of the span extraction question}
\label{appendix:extraction}

\begin{figure}[h]
    \centering
	\includegraphics[width=0.45\textwidth]{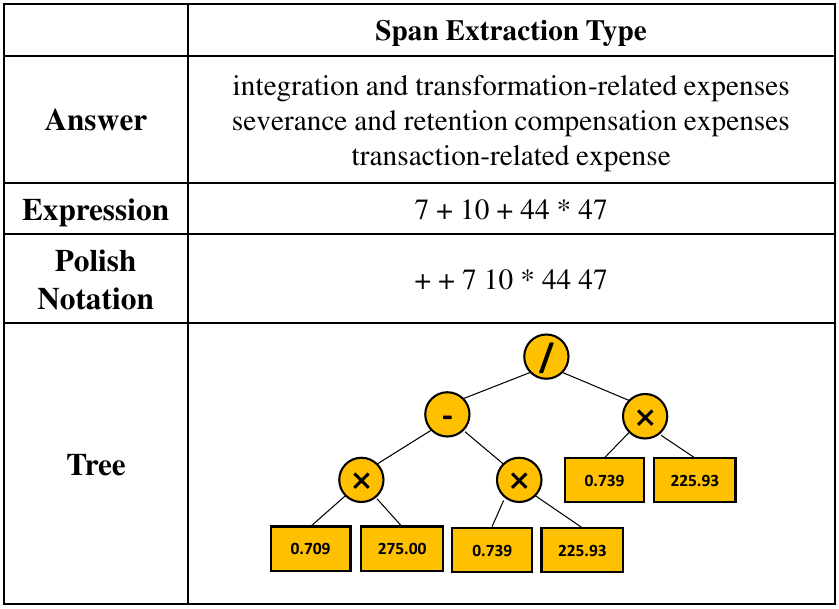}
	\caption{Expressions for span extraction questions. We also predict an expression tree to get the answer. "44×47'' means to select all words between nodes ID 44 and nodes ID 47, ``+'' means that the span on both sides is the answer.}
	\label{fig6}
\end{figure}

We also predict an expression tree for span extraction questions to get the answer. The example is shown in Figure~\ref{fig6}. Unlike mathematical problems, the leaf node in the span extraction tree represents node ID, while the leaf node in the arithmetic tree represents the numeric number or constant. As for the operator, the operators in the arithmetic tree are ``+, -, ×, ÷''. But for the span extraction tree, we define three operators, ``+, ×, C''. ``+'' represents the splicing of two discontinuous spans, corresponding to the \emph{multi-span} in the original dataset. ``×'' means taking the operator's left and right sides as the starting and ending nodes and selecting all nodes in the middle of the two nodes as a span. ``C'' is the same as ``+'', but it counts the number of spans instead of slicing them.

\section{Details of the TAT-QA}
\label{appendix:tatqa}
The specific data analysis of the dataset is shown in Table~\ref{tab:question_detail} and Table~\ref{tab:basic_statistics}.

\begin{table}[h]\small
\centering
\begin{tabular}{@{}lrrrr@{}}
\toprule
           & \textbf{Table} & \textbf{Text} & \textbf{Table-text} & \textbf{Total} \\ \midrule
Span       & 1,801          & 3,496         & 1,842               & 7,139          \\
Spans      & 777            & 258           & 1,037               & 2,072          \\
Counting   & 106            & 5             & 266                 & 377            \\
Arithmetic & 4,747          & 143           & 2,074               & 6,964          \\
Total      & 7,431          & 3,902         & 5,219               & 16,552         \\ \bottomrule
\end{tabular}
\caption{Question distribution regarding different answer types and sources in TAT-QA.}
\label{tab:question_detail}
\end{table}

\begin{table}[h]\small
\centering
\begin{tabular}{@{}lrrr@{}}
\toprule
\textbf{Statistic} & \textbf{Train} & \textbf{Dev} & \textbf{Test}  \\ \midrule
\# of hybrid contexts          & 2,201  & 278   & 278   \\
\# of questions                & 13,215 & 1,668 & 1,669 \\
Avg. rows / table              & 9.4    & 9.7   & 9.3   \\
Avg. cols / table              & 4.0    & 3.9   & 4.0   \\
Avg. paragraphs / table        & 4.8    & 4.9   & 4.6   \\
Avg. paragraph len {[}words{]} & 43.6   & 44.8  & 42.6  \\
Avg. question len {[}words{]}  & 12.5   & 12.4  & 12.4  \\
Avg. answer len {[}words{]}    & 4.1    & 4.1   & 4.3   \\ \bottomrule
\end{tabular}
\caption{Basic statistics of each split in TAT-QA.}
\label{tab:basic_statistics}
\end{table}

\end{document}